\documentclass{article}

\usepackage{CJKutf8}
\usepackage{PRIMEarxiv}

\usepackage{hyperref}       
\usepackage{url}            
\usepackage{booktabs}       
\usepackage{amsfonts}       
\usepackage{nicefrac}       
\usepackage{microtype}      
\usepackage{lipsum}
\usepackage{fancyhdr}       
\usepackage{graphicx}       
\graphicspath{{media/}}     

\newcommand{\zh}[1]{\begin{CJK}{UTF8}{gbsn}#1\end{CJK}}

\title{A RoBERTa-Based Functional Syntax Annotation Model for Chinese Texts}

\author{
  HAN Xiaohui \\
  Faculty of Humanities and Social Sciences \\
  Harbin Institute of Technology \\
  Harbin, China\\
  \texttt{\href{mailto:2007hanxiaohui@sina.com}{2007hanxiaohui@sina.com} } \\
  \And
  ZHANG Yunlong\\
  Faculty of Humanities and Social Sciences \\
  Harbin Institute of Technology \\
  Harbin, China \\
  \texttt {\href{bdvgc@hotmail.com}{bdvgc@hotmail.com}}
   \And
  GUO Yuxi \\
  Department of Computer Science and Engineering \\
  Chalmers University of Technology \\
  Gothenburg, Sweden\\
  \texttt{\href{gggchloe1010@gmail.com}{gggchloe1010@gmail.com}} 
}

\begin{document}
\maketitle

\begin{abstract}
Systemic Functional Grammar and its branch, Cardiff Grammar, have been widely applied to discourse analysis, semantic function research, and other tasks across various languages and texts. However, an automatic annotation system based on this theory for Chinese texts has not yet been developed, which significantly constrains the application and promotion of relevant theories. To fill this gap, this research introduces a functional syntax annotation model for Chinese based on RoBERTa (Robustly Optimized BERT Pretraining Approach). The study randomly selected 4,100 sentences from the People's Daily 2014 corpus and annotated them according to functional syntax theory to establish a dataset for training. The study then fine-tuned the RoBERTa-Chinese wwm-ext model based on the dataset to implement the named entity recognition task, achieving an F1 score of 0.852 on the test set that significantly outperforms other comparative models. The model demonstrated excellent performance in identifying core syntactic elements such as Subject (S), Main Verb (M), and Complement (C). Nevertheless, there remains room for improvement in recognizing entities with imbalanced label samples. As the first integration of functional syntax with attention-based NLP models, this research provides a new method for automated Chinese functional syntax analysis and lays a solid foundation for subsequent studies.
\end{abstract}

\keywords{Functional Syntax  \and Cardiff Grammar \and Systemic Functional Linguistics \and Named Entity Recognition  \and RoBERTa  }

\section{Introduction}
Systematic functional (SF) grammar and transformational generative (TG) grammar exist as two of the most influential theoretical frameworks of modern linguistics. The former, seeing language as a meaning potential, focuses on the social and cultural functions language performs in actual conversational interactions, whereas the latter sees language as a system of rules and a competence of "knowing", focusing on the forms and structures of language and using rules for phrasal structures and transformation to interpret how deep-level structures manifest as surface-level utterance \cite{halliday_introduction_2014} \cite{rong_combining_2017}\cite{huang_cardiff_2023}. Both frameworks should, in theory, complement each other in what they seek to interpret. Yet in reality, while the discipline has seen wide applications of TG grammar in most natural language processing (NLP) systems, serving as the foundation for textual analysis and generation tasks, it has scarcely encountered any NLP systems integrated with or focusing specifically on SF grammar and the communicative functions that texts achieve. This gap signals both a lack of reliable instruments dedicated to SF grammar annotation (with existing ones mostly working on a rule-based system) and impedes the scaled development of functional grammar studies, with researchers having to rely on manual annotation to perform the analysis. 

Among the many branches of systemic functional grammar, Cardiff Grammar (CG) is an important framework that emphasizes the cognitive and interactive nature of language \cite{huang_cardiff_2023}. CG, building upon the foundational works of Halliday, adopts a semantics-centered approach that establishes a clear distinction between meaning and form, treating systems such as transitivity, mood, and theme from the Sydney Model as the meaning potential of language \cite{fawcett_cognitive_1980}\cite{he_representations_2009} \cite{bartley_putting_2018}. Functional syntax is one of the central components of Cardiff Grammar that focuses on the relationship between form (syntax) and semantics (Huang, 2023). It operates through three basic syntactic categories: class of unit, element of structure, and item, which are interconnected through different types of relationships \cite{fawcett_theory_2000}\cite{he_study_2014}. Within this framework, the clause is the core and primary unit comprising multiple structural elements, including Subject, Main Verb, Operator, Binder, and others \cite{zhao_review_2019}.

The fundamental distinction between functional syntax and traditional syntax lies in their approach to analyzing sentence components. While traditional descriptive grammar categorizes sentence elements based on the grammatical functions of words or phrases, functional syntax divides structural components according to their semantic roles, with each structural element corresponding to a specific semantic function. Consider the sentence \zh{我慢慢地把书放在桌子上} (I slowly put the book on the table). Through transitivity analysis, we can identify that the main verb \zh{放} (put) expresses an Action Process. Within this structure, \zh{我} (I) functions simultaneously as the clause's grammatical subject and as the Agent in the action process; \zh{书} (book) serves as both the clause's grammatical complement and the Affected participant in the action; \zh{慢慢地} (slowly) operates as both an adjunct and indicates the Manner of the action process. This phenomenon, where a single expression fulfills different syntactic and semantic roles simultaneously, is termed "conflation" \cite{he_modern_2015}. CG, through this kind of correspondence, establishes how different components within a clause interact with each other and jointly manifest the clause's semantic meaning.

Currently, there is an apparent lack of dedicated NLP models based on the syntactic and functional frameworks of CG, particularly ones focusing on Chinese texts and integrating attention mechanisms with pre-trained models. Development of NLP models based on principles of functional grammar dates back to the 1980s, with early systems like NIGEL \cite{matthiessen_systemic_1983}, WAG (Workbench for Analysis and Generation) \cite{odonnell_input_1996}, and the COMMUNAL Project \cite{fawcett_demonstration_1990} all developing text generation systems grounded in systemic functional grammar. However, these systems relied on rule-based approaches rather than pre-annotated texts and deep learning methods, focusing on text generation and dialogue rather than corpus annotation and analysis. For Chinese specifically, only one computerized tool for manual annotation and result visualization has been developed for functional syntactic analysis \cite{zhang_20_2016}, while attention mechanism-based automatic annotation models remain absent. This gap, among many other factors, led to a significant lag in the scaled promotion of Chinese linguistic analysis. Despite having a history of over two decades and a large research community, applied research accounts for only 4\% of total domestic Cardiff Grammar studies, which is significantly lower than theoretical introduction and exploration (43\%), English syntactic analysis (28\%), and Chinese syntactic analysis (25\%) \cite{zhang_20_2016}\cite{xiang_cardiff_2021}.

Therefore, this study, using the functional syntax framework of CG and framing the identification of structural components as a named entity recognition (NER) task, established one of the first annotated corpora for Chinese functional syntax analysis. Using the annotated corpus, the study trained a RoBERTa-based automatic annotation model for Chinese functional syntax. Results and comparisons with other models demonstrate that the RoBERTa-based model performs well even when trained on a relatively small annotated dataset, making it suitable as a foundation model for further expansion and development.

\section{Related Works}
Functional syntax under the CG framework has been extensively applied in the analysis of Chinese sentence and phrase structures. For instance, Deng \cite{deng_existential_2008} conducted a functional syntactic analysis of existential sentences and their textual functions; He et al. \cite{he_functional_2013} provided a comprehensive description of the syntactic functions of “\zh{要}” (yao); He and Hong \cite{he_study_2014} investigated the semantic and syntactic functions of “\zh{的}” (de) across different sentential contexts. Over the past decade, scholars such as He and Hong have approached modern Chinese from a Cardiff Grammar perspective, using functional syntax to study verb-complement structures \cite{he_study_2014}. Similarly, Zhao and He \cite{zhao_functional_2019} examined modern Chinese “\zh{被}” (bei) constructions through the lens of Cardiff Grammar within systemic functional linguistics, defining the word class of “\zh{被}” and clarifying the meanings expressed by “\zh{被}” constructions in different contexts.

However, as Zhang and Wang \cite{zhang_20_2016} pointed out, research focusing on the actual application of CG, particularly in functional syntax, remains limited, with insufficient integration with other fields. Despite being a linguistic theory that has not only attracted attention but was also successfully implemented in computational language generation tasks (such as the COMMUNAL Project), its corresponding computational implementations and automated text analysis systems have gradually faded from the discipline’s interest, which is surprising considering how rapidly NLP systems have developed in recent years.

Meanwhile, the Transformer architecture, first proposed by Vaswani et al. \cite{vaswani_attention_2017}, introduced the self-attention mechanism that effectively captures long-range dependencies in sequences while enabling parallel training and efficient sequence processing. This architecture has demonstrated significant advantages across various NLP tasks, leading researchers to develop multiple pre-trained language models built on the Transformer foundation. Using Transformer as a basis, Devlin et al. \cite{devlin_bert_2019} introduced BERT (Bidirectional Encoder Representations from Transformers), which employs two pre-training tasks: masked language modeling and next sentence prediction. Through pre-training on large-scale unlabeled corpora, BERT learns deep bidirectional language representations that simultaneously consider both left and right contextual information. Building on BERT’s foundation, Liu et al. \cite{liu_roberta_2019} proposed RoBERTa (Robustly Optimized BERT Pretraining Approach), which systematically optimizes BERT’s pre-training process through various optimization strategies, outperforming other contemporary models across multiple benchmark tests, including GLUE, RACE, and SQuAD.

Compared to English NER, Chinese NER faces additional challenges, including semantic complexity, uncertain entity boundaries, and issues with nested entity relationships \cite{zhang_cross-domain_2024}. To address these challenges, researchers have proposed various BERT-based improvements, such as enhancing BERT’s representation capabilities by incorporating lexical information \cite{zhang_chinese_2018}. Liu et al. \cite{liu_lexicon_2021} introduced the LEBERT model, which directly integrates external lexical knowledge into BERT layers through BERT adapters, significantly improving performance on Chinese sequence labeling tasks. Whole Word Masking (WWM) represents another important method for improving Chinese NER task performance. Cui et al. \cite{cui_pre-training_2021} applied WWM to Chinese BERT pre-training, addressing the problem of words being split into multiple characters after tokenization and improving both accuracy and F1 scores on NER tasks.

\section{The Corpus}
\subsection{Base Corpus Selection}
This study used sentences from the 2014 People’s Daily corpus for functional syntax annotation. The original corpus is an annotated dataset that includes word segmentation and part-of-speech tagging for news articles and commentaries from People’s Daily, which is widely used in various Chinese NER tasks. For this study, 4,100 sentences were randomly selected from the corpus and annotated according to functional syntax theory.

After downloading and selecting the target corpus for annotation, the study cleaned the data by removing existing tags, duplicate sentences, and spaces that might be misidentified as characters. The processed text was then randomly divided into 41 groups, with each group containing 100 sentences of varying lengths. All sentences were converted to JSON format and imported into Label Studio for the actual annotation work.

\subsection{Corpus Annotation}
The corpus annotation was carried out by two annotators with professional backgrounds in linguistics. After receiving training in relevant theoretical frameworks of CG and confirming on annotation guidelines, the annotators proceeded to annotate the target corpus. Based on Fawcett (2000), Huang et al. (2008), and relevant studies on functional syntactic structures of Chinese (e.g., Zhao\& He, 2019), the annotation was carried out in a BIO (B-Beginning, I-Inside, O-Outside) format with the following labels listed in Table 1 to construct the dataset.

\begin{table} [h]
    \centering
\caption{BIO labels and corresponding structural elements  }
\label{tab:my_label}
    \begin{tabular}{ll}\toprule
        \textbf{Label}&  \textbf{Structural Element}\\\midrule
         B-S/I-S&  Subject
\\
         B-M/I-M&  Main Verb
\\
         B-C/I-C&  Complement
\\
         B-A/I-A&  Adjunct
\\
         B-O/I-O&  Operator
\\
 B-X/I-X&Auxiliary
\\
 B-B/I-B&Binder
\\
 B-N/I-N&Negator
\\ \bottomrule
    \end{tabular}

\end{table}

Compared to the original functional syntactic annotation system, this study removed several structural element labels for the following reasons: (1) Certain tags (e.g., concession) existed with extremely low frequency in the corpus; their inclusion would result in severe data distribution imbalance, thereby degrading the performance of the trained model; (2) Certain tags (e.g., infinitives) are not fully compatible with Chinese linguistic conventions; (3) Certain tags (such as main verb extensions) do not adequately correspond to semantic functions like transitivity processes and mood. Therefore, the research team decided to reclassify these tags according to their functional definitions and incorporate them into other tag categories that better suit Chinese linguistic characteristics, thus improving both model applicability and annotation consistency. An example of the annotated sentence is illustrated in Figure 1.

\begin{figure} [h]
    \centering
    \includegraphics[width=0.75\linewidth]{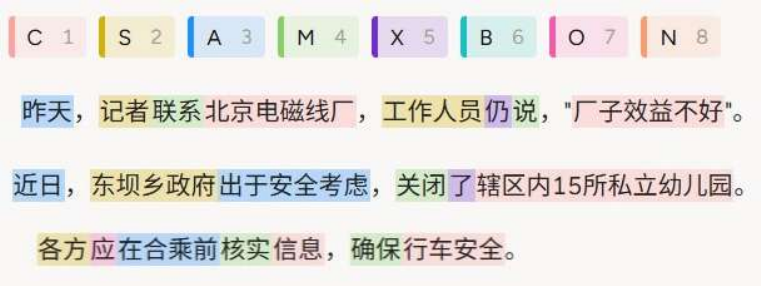}
    \caption{Annotation Example}
    \label{Annotation Example}
\end{figure}

The annotation procedure included pre-annotation and formal annotation. During the pre-annotation phase, the study randomly assigned 400 sentences from 4 groups to each annotator, who annotated the same sentences separately. Upon completion, the team manually checked the annotation results for each sentence. For mismatches, the annotators engaged in discussions with another professional who did not participate in the annotation to clarify and refine the annotation guidelines. During the formal annotation phase, the entire corpus was annotated following the revised guidelines. After completing the corpus-level annotation, the team randomly selected 100 annotated sentences to check overall annotation consistency, using Krippendorff’s coefficient and F1 scores as inter-rater reliability metrics. Results are presented in Table 2.

\begin{table} [h]
    \centering
\caption{Inter-rater reliability results}
\label{tab:my_label}
    \begin{tabular}{lcccccccc}\toprule
        &  S&  M& C& A& O& X& B&N
\\\midrule
         F1&  0.814&  0.880& 	0.867& 0.869& 0.801& 0.836& 0.853&0.853
\\
         $\alpha$&  0.950&  0.957& 0.836& 0.946& 0.876& 0.931& 0.953&0.864
\\
         Support&  217&  	399& 318& 272& 43& 77& 72&24
\\ \bottomrule
    \end{tabular}

\end{table}

Results indicate that all included labels have both F1 score and Krippendorff’s $\alpha$ exceeding 0.8, indicating high annotation consistency and inter-rater reliability. 

After annotation, the corpus was exported for splitting and loading. The dataset used in model training comprised 4,100 samples and 8 label categories in total, and was divided into training (3,280 samples), validation (410 samples), and test sets (410 samples) using an 80\%/10\%/10\% split ratio as shown in Table 3. 

\begin{table} [h]
    \centering
\caption{Label distribution in the dataset}
\label{tab:my_label}
    \begin{tabular}{lcccc}\toprule
        Label&  Train&  Validation& Test&Total
\\\midrule
         M&  5707&  725& 	742&7174
\\
         C&  4828&  610& 647&6085
\\
         S&  3754&  	483& 483&4720
\\
         A&  3559&  434& 417&4410
\\
         X&  1430&  202& 169&1801
\\
 B& 1223& 159& 144&1526
\\
 O& 628& 86& 92&806
\\
 N& 333& 42& 40&415
\\ \bottomrule
    \end{tabular}

\end{table}

All sets included the complete range of labels, with relative label frequencies remaining consistent across sets. For the O and N label categories, which had relatively fewer instances, the study implemented oversampling techniques during fine-tuning to mitigate the effects of label distribution imbalance on training outcomes. 

\section{The Model}
\subsection{Model Selection and Training Strategy}
The study reframed the functional syntactic annotation as a named entity recognition task, treating different structural components as distinct entity types and using the BIO format to handle entity boundaries.

The study selected RoBERTa-Chinese-wwm as the foundation model for fine-tuning, adding a classification head for token classification on top of the pre-trained model using the BertForTokenClassification architecture from Hugging Face’s transformers library. The number of labels and the correspondence between labels and their IDs were specified during model configuration. 

The overall training process includes the following components:

\begin{enumerate}
    \item \textbf{Data preparation}, including data loading and preprocessing, three-way dataset partitioning (training, validation, and test sets in an 8:1:1 ratio), oversampling for entities with smaller sample sizes, and label alignment.
    \item \textbf{Model architecture configuration}, including freezing the first 6 out of the total 12 layers of the model parameters, setting dropout rate as a hyperparameter to be optimized, and replacing the default classification head with a lightweight linear layer to reduce risks of overfitting.
    \item \textbf{Training parameter configuration and baseline training}, employing AdamW optimizer, linear learning rate scheduling with warm-up strategy, calculating class weights based on inverse label frequency to address label imbalance, and implementing early stopping mechanisms to prevent overfitting. The baseline training phase used initial parameters (learning rate $2e^{-5}$, batch size 32, 10 training epochs) for model fine-tuning, monitoring overfitting through a small training subset (20\% of the training set) to provide benchmarks for hyperparameter optimization.
    \item \textbf{Hyperparameter search}, using the Optuna framework to optimize learning rate, batch size, weight decay, and dropout rate, and employing TPE sampler and median pruner to improve search efficiency. The optimization essentially focuses on improving F1 scores of the validation set and reducing overfitting metrics (gap between training set F1 and validation set F1). Each optimization trial conducted 3 rounds of rapid evaluation across three dimensions: validation set F1 (primary metric), training set F1, and overfitting gap (auxiliary metrics).
    \item \textbf{Final training and evaluation}, conducting complete training with optimal hyperparameters, performing comprehensive evaluation and error analysis on the test set, generating reports including loss curves and F1 curves.
\end{enumerate}

Model training relied on PyTorch as the deep learning framework, Hugging Face’s transformers library for model architecture and tokenization functionality, Optuna for hyperparameter optimization, seqeval for entity-level evaluation metrics, and wandb for experiment tracking. Main hyperparameters are presented in Table 4.

\begin{table} [h]
    \centering
\caption{Hyperparameter configuration}
\label{tab:my_label}
    \begin{tabular}{ll}\toprule
        Hyperparameter&  Configuration
\\\midrule
         Learning rate&  $2e^{-5}$ to $5e^{-5}$ (logspace search), wareup ratio 10\%\\
         Batch size&  [16, 32, 64]
\\
         Weight decay&  0.01 – 0.1
\\
         Dropout rate&  0.1\\
         Training epoch&  10 (baseline and final training); 50 (optimization)
\\
 Early-stopping patience&3 epoch
\\
 Freeze layer&6 (58.2
\\ \bottomrule
    \end{tabular}

\end{table}

\subsection{Results}
The study conducted a comparative experiment to evaluate the performance of RoBERTa-Chinese-wwm-ext against several baseline models and methods for NER on the functional syntactic annotation task using the same optimization strategy. The comparative analysis employed four evaluation metrics: precision, recall, F1 score, and overfitting gap to enable cross-model performance comparison (where the overfitting gap was calculated as: overfitting = train F1 - test F1). Results were computed and tracked using seqeval and sklearn libraries, with tracking and aggregation performed on the Weights \& Biases (wandb) platform. The results are presented in Table 5.

\begin{table} [h]
    \centering
\caption{Training result comparison}
\label{tab:my_label}
    \begin{tabular}{lcccc}\toprule
        Model&  Accuracy&  Recall& F1&Overfitting gap
\\\midrule
         CRF+BiLSTM&  0.625&  0.588& 	0.607&0.257
\\
         Transformer (vanilla)&  0.165&  0.256& 0.201&0.024
\\
         BERT-Base-Multilingual&  0.801&  	0.834& 0.817&0.155
\\
         BERT-Base-Chinese&  0.825&  0.847& 0.835&0.165
\\
         RoBERTa-Chinese-wwm-ext&  0.838&  0.866& 0.852&0.129
\\ \bottomrule
    \end{tabular}

\end{table}

Results indicate that the RoBERTa-Chinese-wwm-ext model adopted in this study exhibits superior generalization capability while preserving high accuracy, achieving optimal performance across all evaluation metrics, attaining a precision of 0.838, a recall of 0.866, and an F1 score of 0.852, while maintaining a relatively low overfitting gap of 0.129. The BERT-Base Chinese and BERT-Base-Multilingual models also achieved commendable performance, with F1 scores of 0.835 and 0.817, respectively. Although the traditional CRF+BiLSTM model demonstrated reasonable effectiveness on this task (F1 score of 0.607), its performance was significantly inferior to that of the pre-trained models, with an overfitting gap reaching 0.257. The Vanilla Transformer model exhibited the poorest performance on this task, achieving only an F1 score of 0.201. 

Figure 2 illustrates the evolution of loss curves and F1 curves during the training process.

\begin{figure} [h]
    \centering
    \includegraphics[width=0.8\linewidth]{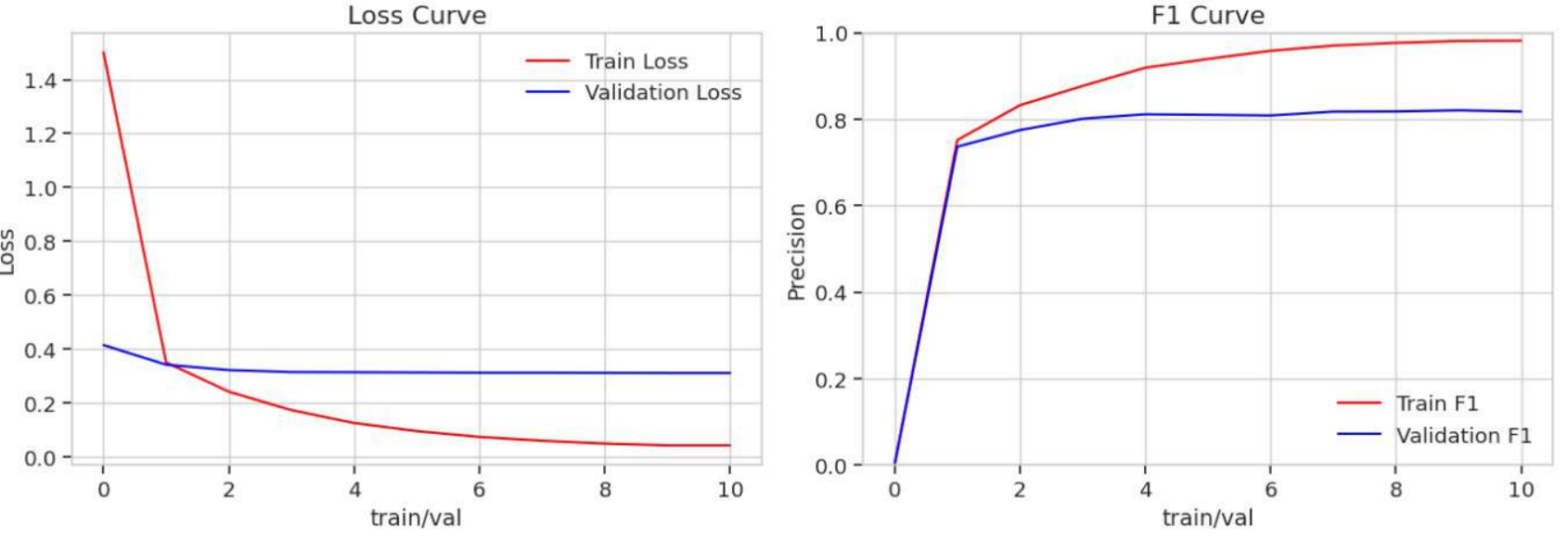}
    \caption{Loss and F1 curve during training }
    \label{fig:placeholder}
\end{figure}

Training loss decreased rapidly from an initial value of 1.5019 to 0.0412 after 10 epochs of training, whereas validation loss exhibited a more modest decline from 0.4141 to 0.3101, plateauing after the 3rd epoch. Meanwhile, the training set F1 score rapidly increased from 0 to 0.9812, while the validation set F1 score rose to 0.8176; growth rates of both training and validation F1 significantly decelerated after the 2nd epoch and exhibited minor variations between epochs 8-10, indicating that the model has approached its performance ceiling under the current data scale. Due to label imbalance and the limited sample size, the model exhibits some degree of overfitting, but this should not be considered severe given the complexity of the annotation task (involving 8 entities and 17 labels). Future work should thus enhance model performance and generalization capability through corpus expansion, vocabulary augmentation, and increasing the diversity of annotated data.

\subsection{Error and Application Analysis}
Figure 3 illustrates label confusion patterns during entity recognition in the final training process and the model's performance on each entity type (measured by F1 score). 

\begin{figure} [h]
    \centering
    \includegraphics[width=0.8\linewidth]{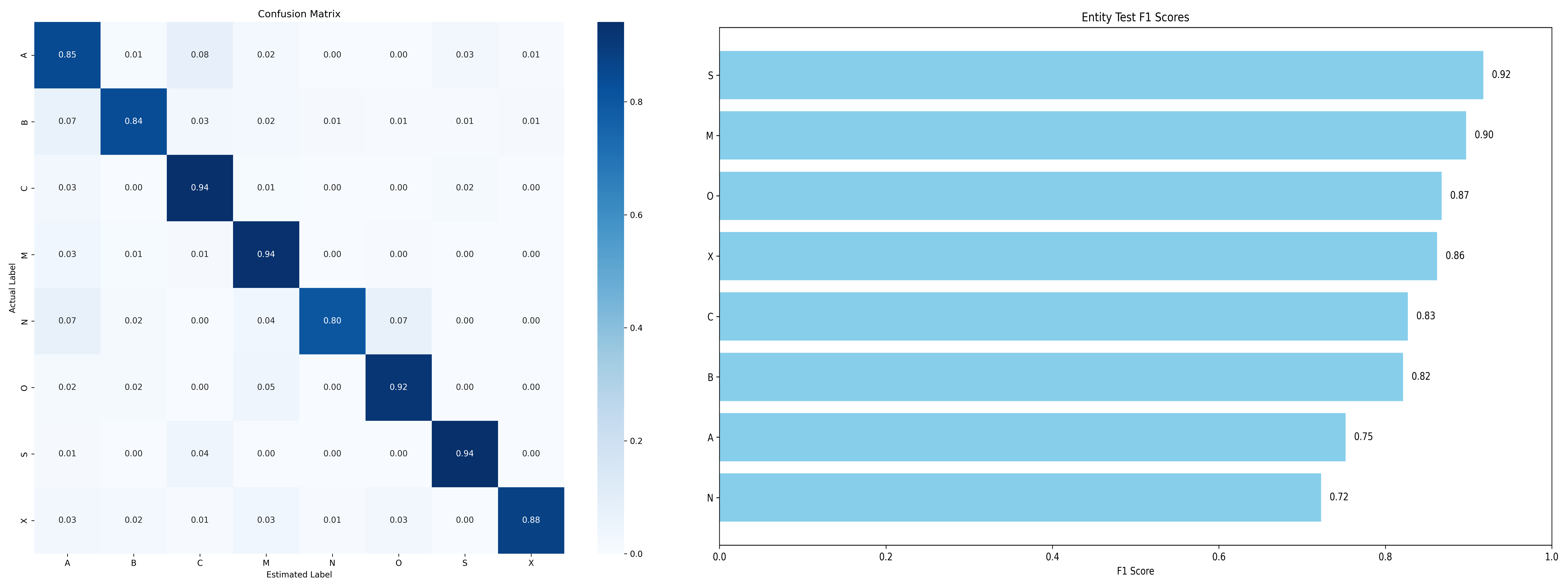}
    \caption{Confusion matrix and entity-level F1 scores}
    \label{fig:placeholder}
\end{figure}

Based on the test set F1 scores, Subject (S) and Main Verb (M) achieved optimal performance due to larger training samples, with F1 scores of 0.918 and 0.897, respectively. The confusion matrix shows that both entity types achieved 94\% correct classification rates with minimal errors, indicating the model's strong capability in identifying core sentence constituents. Operator (O) and Auxiliary Verb (X) achieved F1 scores of 0.868 and 0.862, respectively. The confusion matrix reveals that 92\% of operators were correctly classified, though 5\% were misclassified as Main Verbs (M). This phenomenon occurs because functional syntactic principles allow the same unit in a sentence to be annotated as multiple structural constituents: in some sentences where only an operator exists without a main verb, that constituent can be viewed as both an operator and a main verb (Fawcett, 2000; Huang et al., 2008). For instance:

\begin{itemize}
    \item \zh{他们} [S] \zh{是} [O/M] \zh{记者} [C]
    \item They [S] are [O/M] journalists [C]
\end{itemize}

Here, “are” functions both as a main verb and an operator, corresponding to the “Process” in transitivity (in this case, Relational) while also expressing modal and polarity functions.

Complement (C) and Binding element (B) achieved F1 scores of 0.827 and 0.821, respectively. Complements, with a larger sample size (649), achieved 94\% correct classification with low misclassification rates. Binding elements achieved 84\% correct classification, with 7\% misclassified as Adjuncts (A), indicating some difficulty for the model in distinguishing between binding elements and adjuncts. Adjunct (A) and Negator (N) showed relatively weaker recognition performance. The confusion matrix shows that 85\% of adjuncts were correctly classified, but 8\% were misclassified as Complements (C). Negators, representing the smallest sample category (40 samples), had 7\% misclassified as Operators (O) and 7\% as Binding elements (B). 

A typical annotation error is shown in Figure 4:

\begin{figure} [h]
    \centering
    \includegraphics[width=0.75\linewidth]{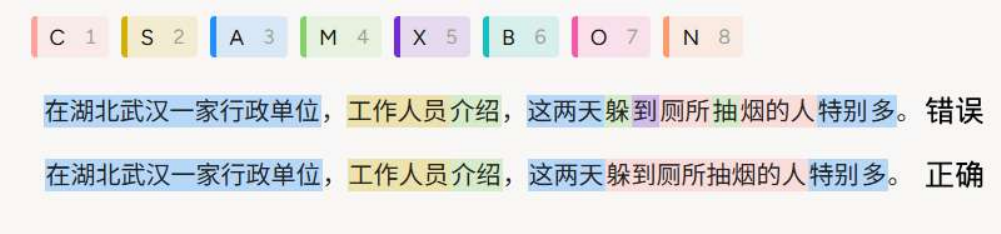}
    \caption{Example of annotation error}
    \label{fig:placeholder}
\end{figure}

In this sentence, the phrase “\zh{躲到厕所抽烟的人} (people who hide in the bathroom to smoke)” constitutes a single complement element that, according to transitivity analysis, functions as the carrier in a relational process and is modified by the subsequent attribute “\zh{特别多} (especially many)”. However, the model divided the different characters within this element and annotated each one, disrupting the correct boundaries of the component. Furthermore, the confusion matrix reveals that although the model demonstrates excellent performance in recognizing core syntactic constituents (Subject, Main Verb, etc.), it shows somewhat inadequate performance when processing Adjuncts and Negators, which are also labeled with the fewest samples in the dataset. 

Primary applications of this model include the investigation of textual microstructures by conducting further annotation based on functional syntactic constituent labeling results, and automatic analysis of extensive texts to examine characteristics of language use at the functional syntactic level. As part of the application experiments, the study randomly selected 190 sentences from the NLPIR Weibo Content Corpus and the People's Daily 2024 Corpus for text cleaning and functional syntactic annotation. 

Table 6 presents functional syntax labeling results using the trained model in analyzing two different types of corpus texts. 

\begin{table} [h]
    \centering
\caption{Functional syntax labeling result comparison}
\label{tab:my_label}
    \begin{tabular}{lcc}\toprule
        \textbf{Text/Label}&  \textbf{Weibo Corpus}&  \textbf{People’s Daily Corpus}\\\midrule
         Total Sentence Count&  190&  190
\\
         Total Word Count&  5747&  8649
\\
         S (Subject)&  271&  	283
\\
         M (Main Verb)&  603&  525
\\
         C (Complement)&  417&  528
\\
 A (Adjunct)& 471&417
\\
 X (Auxiliary)& 317&84
\\
 O (Operator)& 274&31
\\
 B (Binder)& 93&12
\\
 N (Negator)& 89&20
\\ \bottomrule
    \end{tabular}

\end{table}

Comparative analysis of the data reveals that, with equal sentence counts, the Weibo corpus contains significantly more Auxiliary Verbs (X), Operators (O), Binding elements (B), and Negators (N) than the news corpus. The differences are particularly pronounced for Auxiliary Verbs (X) (317:84) and Operators (O) (274:31), indicating that Weibo language tends to favor elements exhibiting strong colloquial and subjective expressions. The high frequency of Negators (N) in the Weibo corpus (89:20) may reflect more diversified subjective attitude expressions in social media contexts. Nevertheless, both corpora show roughly consistent distributions in core sentence constituents: Subject (S), Main Verb (M), and Complement (C), demonstrating linguistic commonalities across different domains. 

These results are, nevertheless, somewhat simplistic and cannot be independently applied to corpus-driven text analysis. Future research should integrate other analytical approaches, such as transitivity, theme-rheme, mood, and voice analyses, to further explore textual characteristics in depth.

\section{Conclusion}

This study proposes an annotated corpus based on Cardiff Grammar’s functional syntax framework and a RoBERTa-based functional syntactic annotation model by transforming the sequence labeling task of functional syntax into an NER task. . While the research has several limitations, including a relatively small sample size, noticeable though not severe overfitting phenomena, and annotation errors on certain entities, it still holds significant theoretical value. As the first integration of functional syntax—and by extension, Cardiff Grammar and Systemic Functional Linguistics—with attention-based pretrained NLP models, the corpus established in this study and the model trained on it effectively fill a gap in related research, enabling Chinese functional syntactic analysis to move beyond manual annotation methods.

Functional syntax serves as a bridging theory between morphology and semantics. Building upon the model proposed in this study, future research can delve deeper into Cardiff Grammar's theory of clausal constituents, analyzing the composition of words and phrases and the relationships that emerge when structural components form clauses. Further studies could also investigate the correspondence between different structural components and semantic functions such as transitivity, mood, voice, polarity, and theme-rheme structure, using more extensive and representative corpora based on different frameworks of functional grammar, and integrate the overall theoretical groundings into the field of natural language processing.

\section*{Acknowledgment}
This project is completed with the assistance of  National Center for Language Technology and Digital Economy Research (No. GJLX20250002), and is funded by Heilongjiang Language Research Committee Project Construction of an Adaptive Intelligent Chinese Learning Platform for International Students in China (No. G2025Y003).

\section*{Data Availability Statement}

Corpus data, label map, training script, and the fine-tuned model can be found and downloaded from \href{https://huggingface.co/bdvgc/RoBERTA_Functional_Syntax}{this repository}.

Note that the corpus file might experience further expansion and/or modification.

\bibliographystyle{unsrt}  
\bibliography{references}  

\end{document}